\documentclass[AMA,STIX1COL]{WileyNJD-v2}
\usepackage{moreverb}
\usepackage{comment}
\usepackage{todonotes}

\newcommand{\tensX}{\mathcal{X}}
\newcommand{\tensY}{\mathcal{Y}}

\newcommand{\tensB}{\mathcal{B}}
\newcommand{\tensF}{\mathcal{F}}
\newcommand{\tensE}{\mathcal{E}}

\newcommand{\tensD}{\mathcal{D}}

\newcommand{\matX}{\textbf{X}}
\newcommand{\matY}{\textbf{Y}}

\newcommand{\matU}{\textbf{U}}
\newcommand{\matV}{\textbf{V}}

\newcommand{\matI}{\textbf{I}}
\newcommand{\matD}{\textbf{D}}
\newcommand{\matC}{\textbf{C}}

\newcommand{\rank}{\text{rank}} 
\newcommand{\R}{\mathbb{R}}

\newcommand\BibTeX{{\rmfamily B\kern-.05em \textsc{i\kern-.025em b}\kern-.08em
T\kern-.1667em\lower.7ex\hbox{E}\kern-.125emX}}

\articletype{ENBIS 2021 Special Issue Article}%

\received{<day> <Month>, <year>}
\revised{<day> <Month>, <year>}
\accepted{<day> <Month>, <year>}

\begin{document}

\title{A Tensor Based Regression Approach for Human Motion Prediction}

\author[1]{Lorena Gril*}

\author[1]{Philipp Wedenig}

\author[2]{Chris Torkar}

\author[1]{Ulrike Kleb}

\authormark{Gril \textsc{et al}}

\address[1]{\orgdiv{Institute for Economic and Innovation Research}, \orgname{JOANNEUM RESEARCH}, \orgaddress{\state{Graz}, \country{Austria}}}

\address[2]{\orgdiv{Institute for Robotics and Mechatronics}, \orgname{JOANNEUM RESEARCH}, \orgaddress{\state{Klagenfurt}, \country{Austria}}}

\corres{\textbf{E-Mail:} *logril@edu.aau.at; firstname.surname@joanneum.at }

\abstract[Abstract]{Collaborative robotic systems will be a key enabling technology for current and future industrial applications. The main aspect of such applications is to guarantee safety for humans. To detect hazardous situations, current commercially available robotic systems rely on direct physical contact to the co-working person. To further advance this technology, there are multiple efforts to develop predictive capabilities for such systems. Using motion tracking sensors and pose estimation systems  combined with adequate predictive models, potential episodes of hazardous collisions between humans and robots can be predicted. Based on the provided predictive information, the robotic system can avoid physical contact by adjusting speed or position. A potential approach for such systems is to perform human motion prediction with machine learning methods like Artificial Neural Networks. 
In our approach, the motion patterns of past seconds are used to predict future ones by applying a linear Tensor-on-Tensor regression model, selected according to a similarity measure between motion sequences obtained by Dynamic Time Warping. 
For test and validation of our proposed approach, industrial pseudo assembly tasks were recorded with a motion capture system, providing unique traceable Cartesian coordinates $(x, y, z)$ for each human joint.
The prediction of repetitive human motions associated with assembly tasks, whose data vary significantly in length and have highly correlated variables, has been achieved in real time.}


\jnlcitation{\cname{%
\author{L. Gril}, 
\author{P. Wedenig}, 
\author{U. Kleb}, and 
\author{C. Torkar}} (\cyear{2021}), 
\ctitle{A Tensor Based Regression Approach for Human Motion Prediction}, \cjournal{journal}, \cvol{year}.}

\maketitle

\section{Introduction}\label{sec1}
Robots are widely used in various industrial and non-industrial applications. 
Common characteristics for the use of robots are precision, reliability, strength and speed. 
In recent decades, various robotic systems have been used in industry in segregated areas that are inaccessible to humans due to safety requirements and regulations.
Current developments in the robotic automation industry show an increasing interest in coexistence of robots and humans in a joint work area in a repetitive assembling task, e.g. assembly assistance, inspection, collaborative production lines, which is indeed a potential future industrial manufacturing scenario.
Therefore, humans working closely with robots rise some very important safety issues which have to be tackled\cite{Villani2018}. 
Collaborative scenarios require accurate and continuous monitoring of the assembling environment of the robot and the human to ensure safe operation. 
Current systems use various types of advanced sensors to identify hazardous situations and to avoid injuries and damage\cite{Castro2021}.
For instance, robot integral force and torque sensors are used to detect collisions with the robot during operation in order to immediately stop the robot. 
Additionally, tactile and touch sensors are used to further improve a robot’s reactive capability. 
Current research has shown how proximity sensors can detect approaching objects at closer ranges and alter the robot behaviour accordingly in advance \cite{MuhlbacherKarrer2017, Navarro2016}. 
Similar to tactile and proximity based sensors research has been conducted on the application of conventional cameras, RGB-D cameras, time-of-flight cameras and laser scanners to monitor the environment\cite{Torkar2019}. 
However, performing sensor data acquisition, analysis, and fusion is a computationally expensive task, that is typically difficult to accomplish with safety-certified devices. 
Nevertheless, there is enormous potential in camera-based perception of the environment.\\
\newline
By observing and predicting human motions in advance, which vary in speed and trajectory as humans usually are not able to repeat the same movements exactly, the robot can foresee hazardous situations and adjust its behavior according to safety standards\cite{2016Robots}.  
Compared to the current approach of tactile and proximity-based safety mechanisms in such applications, predicting the movements of humans and robots increases the time available for initiating appropriate actions to avoid dangerous situations, e.g. collisions.
Currently, safety is ensured by limiting possible contact situations between humans and robots.
Using prediction strives to reduce the actual contact situation in order to detect emergency stop conditions in advance.
Machine learning and deep neural networks have made great developmental leaps in understanding dangerous situations.
Miseikis et al. \cite{Miseikis2018, Miseikis2018a} 
have shown human joint localization in 3D positions using a single image applying a convolutional neural network. 
Similarly, Cao et. al. \cite{Cao2016} 
have shown, how to extract 2D joint positions of multiple humans from images. 
This provides the ability to capture joint motion in an image sequence in a timely manner.
With the increasing number of available data sets containing a variety of human movements \cite{Ionescu2014}, 
research related to human movements is growing.
Working with human motion data requires modeling methods that can handle high-dimensional and correlated data. 
Due to the possibility of high dimensional modeling and fast computations, we have chosen to use a tensor-based method for the prediction of human motion data in a repetitive assembly process.
Our proposed approach uses a linear tensor regression model, where stepwise models are built based on a reference motion. 
Due to the use of the reference, this method has the great advantage that modeling is feasible even with a small amount of data. 
When predicting new motions, a similarity measure is used to select the model, which is the most adequate based on the comparison between the last observed motion sequence and the reference motion. The long-term goal in further developing these methods is to integrate motion prediction into robotic systems, enabling them to plan their own movements in advance and avoid dangerous occurrences like collisions with humans.\\
\newline
Related work regarding capture systems and human motion modelling is described in section \ref{sec2}. 
Since we have opted for a tensor-based method, related work in this field is also presented. 
In section \ref{sec3} some basic definitions regarding tensor algebra are given and the tensor-on-tensor regression model of E. Lock \cite{Lock2017}, on which our proposed method is based, is described. 
Implementation details of our proposed approach are given in section \ref{sec4}. Two different assembling scenarios were outlined and recorded using the OptiTrack\footnote[1]{Motion capture system: \url{https://optitrack.com}} motion capture system and by applying the proposed approach, the modeling was performed. 
The evaluation of the results is given in section \ref{sec5}.
\section{Related Work} \label{sec2}
The research field of tracking humans around robots, especially when these actors are working together, has grown very rapidly in recent years due to more precise capturing systems as well as faster software. These developments are described in \ref{sec2.1}.
Due to the availability of motion data, different methods for prediction in various settings are described in Section \ref{sec2.2}. 
Furthermore, the field of statistical tensor-based methods is briefly described in Section \ref{sec2.3}.

\subsection{Data Acquisition and Human Pose Estimation} \label{sec2.1}

Methods for detecting 3D human body motion in frames are gaining popularity as they are essential for full automation of tracking.
Most methods developed for human motion analysis use models, as kinematic, planar, or volumetric models, to recognize human body parts from given images  \cite{Maurice2019}.
Additionally, different technologies for human pose estimation are currently available, which can be categorised in body mounted and contactless gait kinematic estimation \cite{Zago2020}. Zago et al. \cite{Zago2020} compared the accuracy of marker-less tracking systems, whereas as reference a marker-based system was taken into account. Marker-less systems are based on a camera system, a body model, the image features used and the algorithms that determine the shape, pose and position of the model itself \cite{Colyer2018}.
It is possible to track without markers via pattern recognition in image processing, e.g. recovering 3D human body pose from single images and monocular image sequences \cite{Agarwala} or estimate body pose configuration from a single depth map \cite{Ye2011}. 
A human contour model having the expressive power of a detailed 3D model and the computational benefits of a simple 2D part-based model was introduced by  Freifeld et al. \cite{Freifeld2010}.  Further work has been focused on decreasing the number of installed cameras by Liu et al. \cite{liu2019feature} who showed 3D pose estimation based on single RGB images. 
Xu et al. \cite{xu2020ghum} and Zanfir et al. \cite{zanfir2020weakly} present a statistical, articulated 3D pipeline for modeling human shapes (volumetric model) within a fully trainable, modular deep learning framework, given high-resolution complete 3D body scans of humans captured in various poses. 
Furthermore, marker-less skeleton-based human pose estimation of multiple actors is possible with OpenPose \cite{Slembrouck2020}.
However, marker-based systems for human motion estimation provide very accurate tracking of the human body joints. 
Researchers, like Canton-Ferrer et al. \cite{CantonFerrer2010} or Schönauer et al. \cite{TUW-204383}, developed marker-based optical human motion capture systems.
A set of distinguishable markers are placed on several human body parts, which is captured by a number of calibrated and synchronized cameras providing a robust skeleton tracking. A focus lies also in the labeling of the markers \cite{Bascones2019}.
The representation of the 3D positions of the labeled body parts can be done in different ways, often they are considered in Cartesian coordinate space or, as a further abstraction, in the joint angle space \cite{Torkar2019}.

\subsection{Human Motion Prediction} \label{sec2.2}

As outlined in \ref{sec2.1}, human motion data can be obtained by marker-less or marker-based tracking systems and represented in a Cartesian coordinate system as well as in the joint angle space, both being convertible into each other. 
Human motion prediction aims at forecasting future human poses based on observed past motion or motion patterns. On the one hand, the predictions might focus on gait characteristics, including gait cycle frequency, joint positions or joint kinematics. On the other hand, predictions of 2D or 3D human trajectories might be of interest. Based on the predictions, anomaly detection can be performed by comparing observed and predicted positions or trajectories. Over the past decades, human motion prediction has gained importance in different areas, such as biomechanics, e.g. for healthcare as well as sports, and robotics, especially human-robot collaboration.
Applications in healthcare focus on gait analysis and prediction, e.g. for designing prosthesis for people having leg amputations, as proposed in \cite{Equbal2021}, who observed gait data of different subjects and designed a feed forward neural network to predict future gait angles. Sports applications are presented, for instance, in \cite{Honda2020} for motion prediction in competitive fencing taking into account the interaction between two players by connecting two recurrent neural networks (RNN). An approach for optimizing the rowing technique of professional athletes by modeling motion capture data using bivariate functional principal component analysis is introduced in Becker et al. \cite{Becker2019}. Examples of motion prediction explicitly for robotics are given in, e.g., Gui et al. \cite{Gui2018}, who introduce a motion generative adversarial network (GAN) to teach a robot predicting human motion by observing human activities, and Liu et al. \cite{Liu2017}. The latter propose a new method for modeling a product assembly task as a sequence of human motions using a Hidden Markov model (HMM), thus providing the base for human motion prediction. Mao et al. \cite{Mao2021} motivate their prediction approach by the observation that human motion has a tendency to repeat itself, also in more complex actions occurring across a longer time period and propose a feed-forward neural network to model motion attention by comparing the last visible sub-sequence with a history of motion sub-sequences. Zhang et al. \cite{Zhang2020} point out that most prediction methods in recent research are based on deep learning (DL), including RNN \cite{Martinez2017}, Long Short-Term Memory (LSTM) \cite{Du2019}, gated recurrent unit (GRU) \cite{Sang2020}, conditional variational autoencoder (CVAE) \cite{Ivanovic2019} and convolutional neural networks (CNN) \cite{Nikhil2018}.
Aiming at the detection of anomalies in human motion some interesting research addresses monitoring and predicting physical fatigue of human workers in an industrial environment, which is also relevant for human-robot collaboration. Baghdadi et al. \cite{Baghdadi2019} apply a nonparametric, retrospective, multivariate change point method to gait characteristics of the workers and compare the observed change points with the workers perceived level of exertion for detecting fatigue related changes of walking pattern. A Support Vector Machine (SVM) for automatically recognizing jerk changes due to physical exertion is proposed in Zhang et al. \cite{Zhang2019}.
\newpage

\subsection{Tensor Based Methods for (Motion) Prediction} \label{sec2.3}
The focus of this paper is on simultaneously predicting the position of certain joints (markers) of a human body for a specified period of time, i.e., forecast horizon. Conventional vector or matrix-based regression models are not able to adequately handle higher-order tensors as these data contain structural information (i.e., information about the correlation patterns between different modalities in higher-order tensors).
Motivated by the wide applicability of tensor data and the limitations of classical regression models, in the sense that the multidimensional structure of the data cannot be properly represented and they are difficult to interpret too, the focus is to use a tensor-based regression model.
The strength of tensor models lies in the decomposition methods that can represent high order data in terms of low dimensional factors, having applications in  a variety of fields
such as chemometrics, image/video analysis, neuroscience  and signal processing\cite{KoBa09}.
Two fundamental decomposition formats are canonical decomposition/parallel factor analysis (CP) \cite{Carroll1970, Harshman1970FoundationsOT} and Tucker \cite{Tuck1963a,Lathauwer2000} decomposition.
Due to the wide applicability of tensor-based models, there is also a large variety of regression models with different requirements with respect to the input and output data type as well as the determination methods\cite{Hou2017}. 
The CP regression developed by Zhou et al.\cite{Zhou2013} and the Tucker regression by Li et al.\cite{Li2013} with applications in neuroimaging, are both scale-on-tensor models, of the form $  y = \langle \tensX, \tensB \rangle + \varepsilon$ with the restriction that the coefficient tensor $\tensB$ needs to have a CP or Tucker structure. 
Tensor-on-tensor models estimate a multidimensional tensor consisting of multiple correlated data from another tensor. 
In the High Order Partial Least Squares regression (HOPLS) model, the data is explained by a sum of orthogonal Tucker tensors, while the number of orthogonal loadings serves as a parameter to control model complexity and prevent over-fitting \cite{Zhao2013}. However, the standard HOPLS can quickly become prohibitively expensive to compute, which is why an online tensor regression algorithm, namely Incremental High Order Partial Least Squares  (IHOPLS), has been developed. A field of application of the IHOPLS is to reconstruct 3D motion trajectories from video and ECG stream signals \cite{Hou2016}. Lock \cite{Lock2017} provides a tensor-on-tensor regression method with a CP structure of the regression parameters using least squares and Llosa \cite{Losa2018} uses the Tucker structure of the coefficients. Gahrooei et al. \cite{Gahrooei2020} propose a functional regression method in which a high dimensional response is estimated and predicted by a set of informative and non-informative high dimensional covariates through a set of low-dimensional smooth basis functions. An application of the proposed functional regression method are joint motion trajectories. Currently, a lot of research is being done on tensor-based methods, including tensor-based networks \cite{Zniyed2021, Wu2021, Wang2020a} and Bayesian approaches \cite{Guhaniyogi2015}.

\section{Description of the Proposed Approach}\label{sec3}
Our proposed approach for predicting human motion data in an industrial repetitive assembling process is based on  tensor-on-tensor regression model introduced by E. Lock. 
Therefore, basic definitions related to tensor algebra are provided in Section \ref{sec3.1} to develop a deeper understanding for our proposed method. A brief description of our approach is given in Section \ref{sec3.2}

\subsection{Theoretical Background} \label{sec3.1}
A \textbf{\textit{tensor}} is defined as a multiway array of real numbers $\tensX \in \mathbb{R}^{I_{1} \times \dots \times I_{D}}$ of order $D$. 
The $d$-mode vector of tensor $\tensX$ is an element of $\R^{I_{d}}$, which is obtained by varying the index $I_{d}$ while keeping other indices fixed.
The $d$-mode unfolding of a tensor $\tensX$, called \textbf{\textit{matricization}}, is the process of rearranging the $d$-mode vectors into the columns of the resulting matrix $\matX_{(d)} \in \R^{I_{d} \times I_{1}\cdot\, \dots\, \cdot I_{d-1} \cdot I_{d+1} \cdot\, \dots\, \cdot I_{D}}$.
The \textbf{\textit{vectorization}} of $\tensX$ is defined as the vectorization of its $1$-mode matrix. 
The tensor $\tensX$ is a \textbf{\textit{rank-one}} tensor, if it can be represented as the outer product of $D$ vectors $\{a^{(d)}\in \R^{I_{d}}\}_{d=1}^{D}$. 
Furthermore, the \textbf{\textit{tensor rank}} of $\tensX$ is defined to be the minimum number of the sum of rank-one tensor that can exactly factorize the tensor $\rank(\tensX) := \mathrm{arg}\min\{R \in \mathbb{N}: \tensX = \sum_{r=1}^{R} a_{r}^{(1)} \circ a_{r}^{(2)} \circ \dots \circ a_{r}^{(D)} \}$. 
Let $\matX \in \R^{I \times J} $ and  $\matY \in \R^{K \times L} $, the \textbf{\textit{Kronecker product}} is defined as $\matX \otimes \matY =  (x_{ij} \matY )_{i = 1, \dots I; \; j = 1, \dots J}$. The \textbf{\textit{Frobenius norm}} of the 
tensor $\tensX$ is defined as the square root of the inner product of $\tensX$ with itself.
For the tensors $\tensD \in \R^{I_1 \times \dots \times I_K \times P_1 \times \dots \times P_L }$ and $\tensF \in \R^{ P_1 \times \dots \times P_L \times Q_1 \times \dots \times Q_M }$ the  \textbf{\textit{contracted tensor product}} along $L$ modes is elementwise defined as $ \langle \tensD, \tensF \rangle_L [i_1, \dots i_K, q_1, \dots q_M ] = \sum_{p_1 = 1}^{P_1} \dots \sum_{p_L = 1}^{P_L} \tensD_{i_1, \dots, i_K, p_1, \dots p_L} \tensF_{p_1, \dots, p_L, q_1, \dots q_M} $.  
The strength of tensor modeling lies in its associated decomposition tools that are capable of representing high-order data in terms of low-dimensional factors.
For the tensor based regression method proposed by Eric Lock the CP decomposition is used.
The  \textbf{\textit{CP decomposition}} of a tensor generalizes the bilinear factor models to multilinear data. 
Mathematically, the R-component CP model factorizes a
tensor $\tensX \in \mathbb{R}^{I_{1} \times \dots \times I_{D}}$ as a linear combination of rank-one tensors 
$\tensX = \sum_{r=1}^{R} \lambda_{r} a_{r}^{(1)} \circ a_{r}^{(2)} \circ \dots \circ a_{r}^{(D)} + \epsilon$ 
with tensor rank $R$, scale weighting of the different rank-one tensor components $\lambda_{r}$ and the residual tensor $\epsilon$ \cite{ Hou2017, Hackbusch2020, Cichocki2009}. \\

Our proposed modeling approach is based on a \textbf{\textit{tensor-on-tensor regression}} of Eric Lock\cite{Lock2017}. 
The contracted tensor product for the linear prediction of an outcome array $\tensY \in \R^{N \times Q_{1}\times \dots \times Q_{M}}$ 
from the predictor array $\tensX \in \R^{N \times P_{1} \times \dots \times P_{L}}$  is defined as
\begin{equation}\label{eq:linreg}
    \tensY = \langle \tensX, \tensB \rangle_{L} + \tensE,
\end{equation}
where $\tensB = [\matU_1, \dots, \matU_L, \matV_1 \dots, \matV_M] \in \R^{P_1 \times \dots \times P_L \times Q_1 \times \dots \times Q_M}$ is the coefficient array, factorized via CP decomposition, and   
$\tensE \in \R^{N \times Q_1 \times \dots \times Q_M}$ is the error array.  
Regarding well-definedness, the multidimensional structure of $\mathcal{X}$ and $\mathcal{Y}$, stability as well as over-fitting, a low-rank solution with 
$L_2$ penalty term is proposed and leads to the objective 
\begin{equation}\label{eq:argmin}
    \mathrm{arg}\min_{\rank\left(\mathcal{B}\right)\le R} \left|\left|\mathcal{Y}-\left\langle\mathcal{X},\mathcal{B}\right\rangle_L\right|\right|_F^2+ \lambda\left|\left|\mathcal{B}\right|\right|_F^2.
\end{equation}
The solutions, with respect to the factorization matrices (for all modes due to permutation invariance of the loss function), are
$$\text{vec}(\matU_1) = (\matC^T \matC + \lambda(\matU_2^T \matU_2 \cdot \dots \cdot \matU_L^T \matU_L \cdot \matV_1^T \matV_1 \cdot \dots \cdot \matV_M^T \matV_M) \otimes \matI_{P_1 \times P_1})^{-1} \matC^T \text{vec}(\tensY)$$
and 
$$\matV_M = (\matD^T \matD + \lambda(\matU_1^T \matU_1 \cdot \dots \cdot \matU_L^T\matU_L \cdot \matV_1^T \matV_1 \cdot \dots \cdot \matV_{M-1}^T \matV_{M-1}))^{-1} \matD^T \matY_M^T,$$ 
where $\matC$ is the concatenation of the matricization of the contracted tensor product of $\tensX$ for all components of $R$, the CP factorization without $\matU_1$ and $\matV$ is analogously defined without considering $\matV_M$. 
To quantify the uncertainty of the resulting point estimation, without assumptions regarding the distribution of the data, a MCMC Gibbs sampling method iteratively updating the variance of entries of $\tensE$ and the factor matrices is proposed.
Uncertainty of the predictions with a $\tilde{N}$ out-of-sample observations with $\tensX_{\text{new}} \in \R^{\tilde{N} \times P_1 \times \dots \times P_L}$
 can be assessed by  
 sampling from the predictive posterior distribution of
\begin{equation}\label{eq:preddist}
    \tensY^{(t)}_{\text{new}} = \langle \tensX_{\text{new}}, \hat{\tensB}^{(t)} \rangle_L + \tensE^{(t)}_{\text{new}}.
\end{equation}

\subsection{Proposed Approach} \label{sec3.2}
To development a tensor-based human motion prediction model, an assembly task performed by a person with multiple repetitions is used. 
Our proposed algorithm requires joint angle data as input and the modeling is performed on a reference motion due to the unequal duration of the repetitions of the assembly task. 
Detailed descriptions of the transformation of the Cartesian coordinates, given by the OptiTrack system, into the joint angle space as well as the determination of the reference cycle can be found in Section \ref{sec:4.2.1}. 
The goal of predicting repetitive assembly tasks with a linear tensor-based regression model is to use the last $l$ of past seconds to predict $k$ of them into the future, $k \leq l$. 
A dense collection of coefficient tensors was generated using the reference, see \ref{sec:4.2.2}. 
When predicting a new assembly process, with same motion pattern but with different length and some unavoidable variation in trajectories, appropriate coefficient tensors have to be selected. The selection is based on Dynamic Time Warping\cite{Giorgino2009} where the past information of the motion of length $l$ is compared to the complete reference motion. By means of the information about the highest similarity, the appropriate coefficient tensor can be selected and the prediction can be performed by using the contracted product. A more precise description of this part can be found in Section \ref{sec:4.2.3}.
In addition, a simulation study was conducted to determine the uncertainty for the predictions, described in Section \ref{sec:4.2.4}.

\section{Experiments}\label{sec4}
In the following chapter, a detailed description of the prediction of human movements in an industry-oriented setting by the outlined tensor-based approach is given.
In Section \ref{sec4.1} a description of the two data sets used is provided and furthermore, in Section \ref{sec:4.2} implementation details of our proposed approach are presented. This section is divided into the segments of pre-processing of the data \ref{sec:4.2.1}, modeling \ref{sec:4.2.2}, prediction \ref{sec:4.2.3}, and uncertainty quantification \ref{sec:4.2.4}. 

\subsection{Data Sets}\label{sec4.1}
For the development and evaluation of the proposed approach we have compared different available human pose and motion data sets such as Human3.6M\cite{Ionescu2014}.
Most available data sets do not focus on assembly tasks in an industrial environment but on sports activities or common activities like walking\cite{Siirtola2018}. To further emphasize the focus on industrial applications we decided to record and prepare two own data sets. In order to use the installed camera and sensor network infrastructure of the company, the recordings of assembly processes in the industrial setting were performed in our laboratory environment \footnote[2]{Laboratory environment used for data acquisition:  \url{https://www.joanneum.at/robotics/infrastruktur/hands-on-area}}. The experiments conducted focused on repetitive assembly and screwing tasks common in automotive and consumer electronics production. Actors were recorded working on a workbench using a camera-based motion capture system, more specifically the OptiTrack system. The actors were equipped with reflective markers attached to certain parts of the torso, namely, on the hip, spine, neck, head, as well as on the left and right shoulder, elbow and hand. In this recording, the 3D positions of the markers on certain body parts were measured in Cartesian space with respect to a predefined zero point using camera images of the system. To obtain 3D coordinates in space at least three cameras are necessary; in our setting we used four OptiTrack cameras placed at different positions. The settings regarding the recorded data sets are summarized in Table \ref{table:tab1}.
\begin{table}[h]
\begin{center}
\centering
\begin{tabular}{lll}
\hline
\hline
\multicolumn{3}{c}{\textbf{Summary of the Recording Details}}                                     \\ \hline \hline & \\[-0.3ex]
\multicolumn{1}{l}{}                             & \multicolumn{1}{l}{\textbf{\textit{Data set 1}}}               & \textbf{\textit{Data set 2}}                           \\ \hline
\multicolumn{1}{|l|}{Recording system}           & \multicolumn{1}{l|}{ Optitrack}                                & \multicolumn{1}{l|}{ Optitrack}                        \\ \hline
\multicolumn{1}{|l|}{Returned data format}       & \multicolumn{1}{l|}{Cartesian coordinates {[}x,y,z{]}}         & \multicolumn{1}{l|}{Cartesian coordinates {[}x,y,z{]}}  \\ \hline
\multicolumn{1}{|l|}{Actions}                    & \multicolumn{1}{l|}{Pseudo assembly process}                   & \multicolumn{1}{l|}{Pseudo screwing process}             \\ \hline
\multicolumn{1}{|l|}{Num. of actors}             & \multicolumn{1}{l|}{5 }                                        & \multicolumn{1}{l|}{1 }                                   \\ \hline
\multicolumn{1}{|l|}{Actor(s) pose}              & \multicolumn{1}{l|}{Standing}                          & \multicolumn{1}{l|}{Standing }                            \\ \hline
\multicolumn{1}{|l|}{Num. of recorded joints }   & \multicolumn{1}{l|}{10}                                        & \multicolumn{1}{l|}{10}                                    \\ \hline
\multicolumn{1}{|l|}{Num. of total frame}        & \multicolumn{1}{l|}{71 993}                                        & \multicolumn{1}{l|}{22 001}                        \\ \hline
\multicolumn{1}{|l|}{Recording length}           & \multicolumn{1}{l|}{39 min 59 sec}                             & \multicolumn{1}{l|}{6 min 7 sec }                       \\ \hline
\multicolumn{1}{|l|}{Frame rate}                 & \multicolumn{1}{l|}{30 Hz}                                     & \multicolumn{1}{l|}{60 Hz }                                \\ \hline

\end{tabular}
\caption{Summary of the Recording Details of our two Data Sets} 
\label{table:tab1}
\end{center}
\end{table}
\subsubsection{Data Set 1: Assembling and Disassembling Task} \label{sec:assembling}
\begin{figure}
    \centering
    \includegraphics[width = 10 cm]{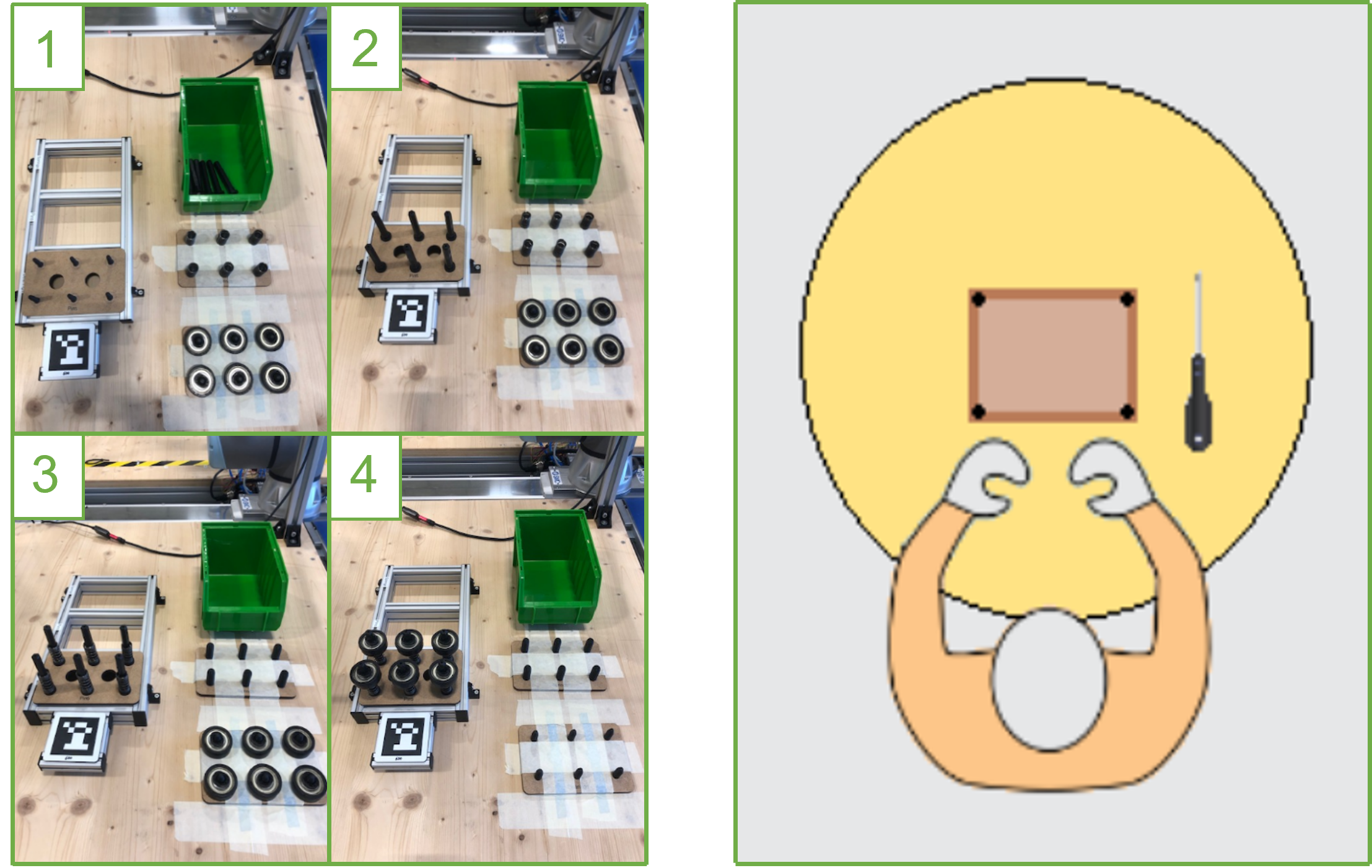}
    \caption{\textit{Visualization of the Assembling and Disassembling Task (left) as well as the Screwing Task (right).} For data set 1, five persons of different physique assemble and disassemble six pins, springs and ball bearings on protruding struts in a specific order; for data set 2, one person screwed 4 screws.}
    \label{fig:Dataset12}
\end{figure}

In order to develop the modeling with the proposed tensor-based method a complex repetitive assembly task executed by several individuals with different physical conditions, was recorded with the OptiTrack system using a frequency rate of 30 Hz.  
Following a predefined scheme, each actor first placed 6 pins on a plate with protruding struts, then a spring and finally a ball bearing was placed on each of them. The described process is referred to as the assembly process and depicted in Figure \ref{fig:Dataset12} (right).  
In order to simulate the disassembly process, this setup was reassembled back to the starting point. 
An assembly and disassembly cycle took about 1 minute and 10 seconds, but depended heavily on the working person. 
\subsubsection{Data Set 2: Screwing Task} \label{sec:screwing}
In order to evaluate and to optimize the model parameters, records of a screw movements, which are supposed to simulate a assembly line scenario, were used. 
As visualized in Figure \ref{fig:Dataset12}, a plate with a screw hole in each corner is placed on a table.
One assembly cycle consists of screwing four screws into the plate in a predefined order. 
For each screw, a screwdriver and a screw were grabbed, the screw was tightened in one of the corners, and then the screwdriver was returned to its rest position.
Markers on joints of the actor's torso were placed and at a frequency of 60 Hz, the movements were recorded using the OptiTrack system \cite{Torkar2019}.

\subsection{Implementation Details} \label{sec:4.2}
The entire implementation was done in the programming language R, and an R-Shiny app was developed for a fast and straightforward data preparation and prediction. 
In order to implement real time predictions of repetitive motions in an industrial assembly process, special attention must be paid to the high dimensionality of the data and the high correlation between individual joints as well as the different duration of the movements of the individual processes and the small amount of data available.

\subsubsection{Pre-processing}\label{sec:4.2.1}
\begin{figure}[h]
    \centering
    \includegraphics[width = 10cm]{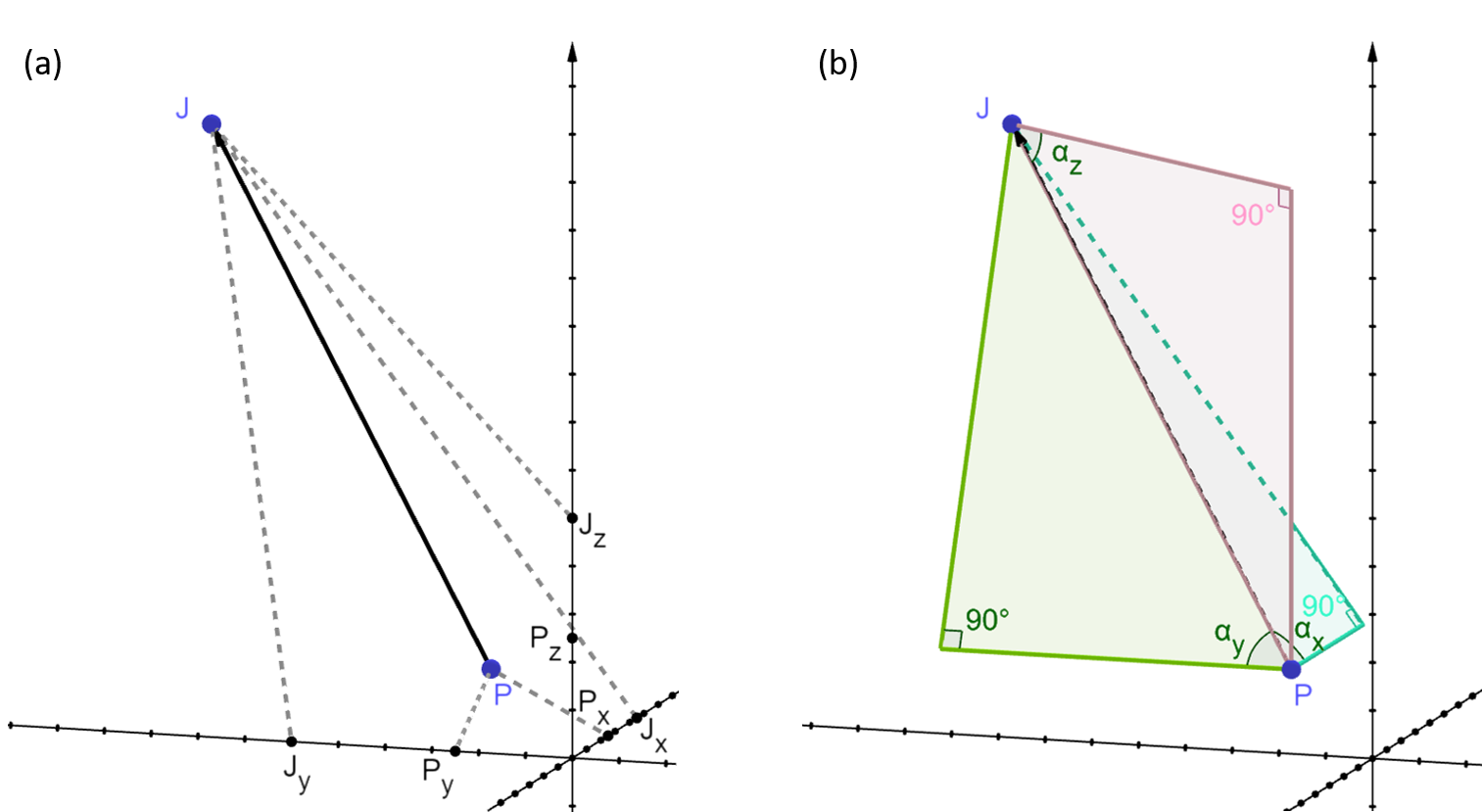}
    \caption{\textit{Illustration of the Transformation of the Cartesian Coordinates into the Joint Angle Space.} (a) The Cartesian coordinates of the joints $J$ and $P$ are shown. (b) Along each axis the cosine theorem according to formula \eqref{eq:CCtoJAS} is applied shown as rectangles with the Euclidean distance between the joints as hypotenuse and the distances between a fixed axis as adjacent.}
    \label{fig:JAS}
\end{figure}

The data acquisition by the OptiTrack system provides the corresponding Cartesian coordinates for each marker placed on certain relevant parts of the actor's body.
For the identification of the markers to the corresponding body parts, a fill-in bar was set up in the app. 
The data acquisition by the OptiTrack system provides the corresponding Cartesian coordinates for each marker placed on certain relevant parts of the actor's body.
For the identification of the markers to the corresponding body parts, a fill-in bar was set up in the app. 
The complexity of the human body forces the community working on human motion prediction taking the proportions of a human body into account and therefore, the collected data is transformed into the joint angle space. 
Because of computational cost, the Cartesian data were not transformed via the classical forward and inverse kinematic model.
In our approach, the Euclidean distance between connected markers was determined, and the angles were computed along each coordinate of the space-fixed Cartesian coordinate system using the cosine theorem. 
In detail, let $J$ be a arbitrary joint and $P$ be its predecessor. With the Euclidean distance $d_{(P, J)}$ between them, the component-wise determination of the angle along all axis $v = \{x, y, z\} $ is 
\begin{equation}\label{eq:CCtoJAS}
    \alpha^{(P,\, J)}_v = \arccos \Biggl(  \frac{J_v - P_v}{d_{(P, \, J)}} \Biggr).
\end{equation} 
Figure \ref{fig:JAS} visualizes the transformation of the Cartesian coordinates of joint $P$ and $J$ (a) to the joint angles (b), where $d$ denotes the Euclidean distance and the colored triangles are intended to illustrate the calculation of the angle along each axis.

Fast Fourier Transform (FFT) was applied to a selected angle of the joint angle space, which best represents the motion of the task, to identify each of the repetitive assembly cycles, resulting in a sufficiently smooth progression over time. 
Since humans are not capable of repeating movements to an accuracy of (milli)seconds, the assembly processes we work with are not of equal length. In the assembling and disassembling task of \ref{sec:assembling} the individual assembly/disassembly cycles are separated after the completion of 18 sub-tasks, representing the 6 pins, 6 springs and 6 ball bearings.  
Furthermore, as described in \ref{sec:screwing}, one screwing task is completed after the occurrence of the $4^{th}$ peak of the transformed data.
For building the tensor-on-tensor regression model, a reference cycle (or reference movement) for one assembly task, consisting of all joint angles, was generated based on few selected cycles of the recorded data. 
After stretching or compressing these cycles to equal length, the arithmetic means of them were computed for each time step, resulting in the reference cycle.

\subsubsection{Modeling}\label{sec:4.2.2}
\begin{figure}[h]
    \centering
    \includegraphics[width = 11cm]{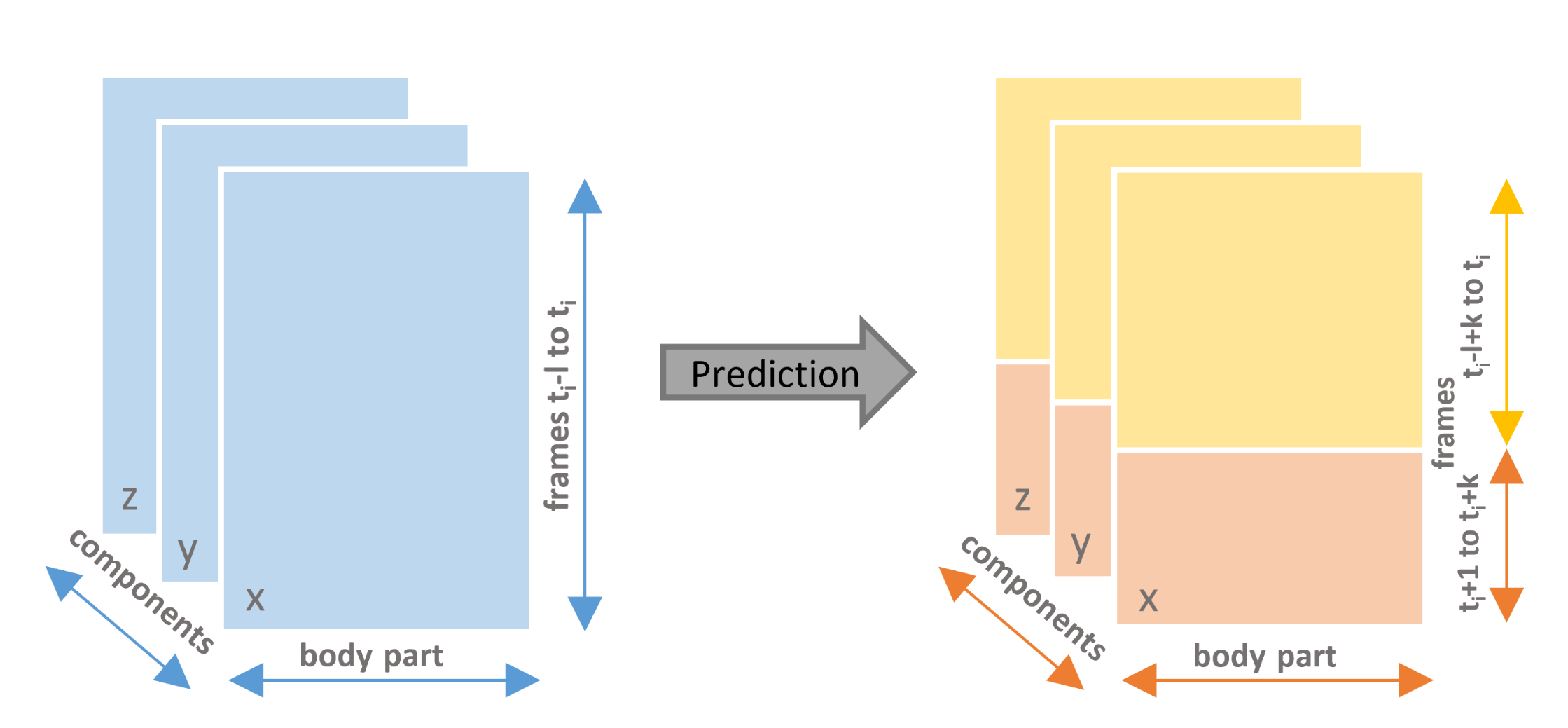}
    \caption{\textit{Representation of the Input and Output of the Modeling Method.} The Input and Output represents $l \times 10 \times 3$ tensors where $l$ represents the number of observations or frames, $10$ the number of joints (body parts) and $3$ the axes (components). The input contains $l$ seconds of past observation and the output $l-k$ seconds of past and $k$ of future information.}
    \label{fig:Prediction}
\end{figure}

\begin{figure}[h]
    \centering
    \includegraphics[width = \textwidth]{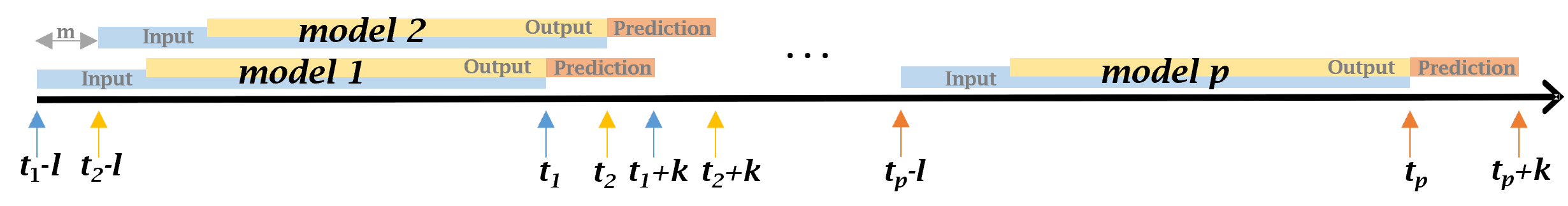}
    \caption{\textit{Illustration of the Iterative Modeling Process.} A prediction model every $m$ seconds is build, denoted by model 1 to p. Therefore, input and output tensors are generated from the data having the same number of observations $l$, so that the input tensor contains past information of $l$ seconds, and the output tensor contains $l-k$ seconds of past information and $k$ seconds of future information.}
    \label{fig:IMP}
\end{figure}

The modeling was performed on the angle-based data of the reference cycle.
Due to the intention to predict $k$ seconds into the future where $l$ seconds of the past are given, an iterative modeling along the reference is necessary. 
The modeling using linear tensor regression takes the knowledge of the past as predictor and the information to be predicted as predictive. Hence, the computation of the corresponding coefficient tensor is essential. 
Because of missing past information the modeling of the first $l$ seconds is not possible. Therefore, the last $l$ seconds of the reference are duplicated and appended to the beginning of the reference. This process is reasonable because of the consideration of repetitive movements. During iterative model building, the respective input and output joint angle data are filtered out from the reference cycle and restructured into a 3D array, where the first dimension represents the number of observations (frames), the second dimension corresponds to the number of markers on the body (i.e. the body parts) and the third dimension reflects the $x$, $y$, and $z$ coordinate axes (i.e. components) along which the angles have been determined. 
As the predictor and the predictive tensor need to have the same number of observations corresponding to the first component, the output tensor must be modified. 
The output tensor is enlarged by past information, i.e. the predictive tensor contains $l-k$ seconds of the past and $k$ seconds of future information ($k \leq l$). The structure of the predictor and the predictive tensor is depicted in Figure \ref{fig:Prediction}.
The estimation of the coefficient tensor is based on the predictor and the outcome array using the \texttt{rrr()} function from the package \texttt{MultiwayRegression} and requires the determination of tuning parameters $R$ and $\lambda$, where $R$ is the rank of the tensor and $\lambda$ describes the $L_2$ penalty term, respectively.
During the iterative modeling process along the extended reference cycle, a coefficient tensor is estimated every $m$ frames and stored in a list along with the angle designation, resulting in a collection of coefficient tensors, referred to as dense $\tensB_{i}$-collection, with $(i = 1, \dots, p)$ . Hence, the modeling of $k$ seconds of human motion data given $l$ seconds of past information every $m$ frames via $p$ overlapping models is visualized in Figure \ref{fig:IMP}. 
\subsubsection{Prediction} \label{sec:4.2.3}
Based on the model developed on the reference cycle, other motions reflecting the same assembly process will henceforth be predicted. Since humans are not able to duplicate movements exactly, the length of these assembly processes are typically different with respect to the reference. To perform a prediction, given knowledge of a sequence of the current motion of length $l$ seconds, a suitable coefficient tensor, that may best represent the future motion, has to be used. As the movements of the person performing the assembling process vary in speed, it is not advisable to select a coefficient tensor in the dense $\tensB_{i}$-collection $(i = 1, \dots, p)$ in an iterative way with constant increments. For the determination of the best coefficient tensor, the current cycle and the extended reference are compared in terms of similarity of motion patterns. 
In order to address the comparison of data varying in speed, Dynamic Time Warping (DTW) is used for measuring similarity between two temporal sequences. 
By using the DTW algorithm the coefficient tensor with the highest similarity with
respect to the extended reference is identified. In determining the optimal coefficient tensor, the match of the last frame of the current cycle with the extended reference is considered and, since a model was only built every $m$ frames, the coefficient tensor $\tensB_{i}$ with time index $t_i$ closest to that match is used for prediction. For the prediction of $\tensY_{new}$ the contracted product between the new input tensor $\tensX_{new}$ and the ideal coefficient tensor $\tensB_{opt}$ is calculated by using the \texttt{ctprod()} function of the \texttt{MultiwayRegression} library. Since past data was also used in the modeling process for the predictive tensor, only the last $k$ seconds of $\tensY_{new}$ correspond to the prediction. In order to exploit all existing information regarding the history of the new motion, the predictions are appended to the last observed position data.
Due to the fast calculations of $\tensB_{opt}$ and the contracted product, additional updates of the predictions are possible after $u \leq k $ seconds. Hence, when predicting new motions of an assembling process, the inputs needed are the new data sequences, the extended reference, the number of frames $m$ representing the density of the $\tensB_{i}$-collection, the number of seconds of the predictor and the predictive as well as the number of frames $u$ corresponding to the next update. 
\subsubsection{Uncertainty Quantification} \label{sec:4.2.4}
The goal is to assess uncertainty for the predicted movements of the actor which was done by determining 
the predictive variation on the one hand, 
and the posterior predictive distribution with the method proposed in \cite{Lock2017} on the other hand, both based on the reference cycle. With these two methods we get predictive uncertainty intervals by using the standard deviation of prediction. \\

\textbf{Predictive Variation: } The predictive variation is used to describe the variations of the prediction results on each of the $p$ linear prediction models. We chose to use an ensemble approach, which trains on multiple copies of the model\cite{Yu2021}. 
To evaluate the predictive variation, we generated 1000 samples of the reference cycle by adding a Standard Normal random number, multiplied by a specified variability, to the original reference cycle. This variability was determined from the observations of the individual stretched or compressed cycles that formed the reference cycle at each instant of time. The predictive variation of the p models was computed by generating predictions for each of the sampled cycles using the respective model based on formula \eqref{eq:linreg}. 
\begin{figure}[h]
    \centering
    \includegraphics{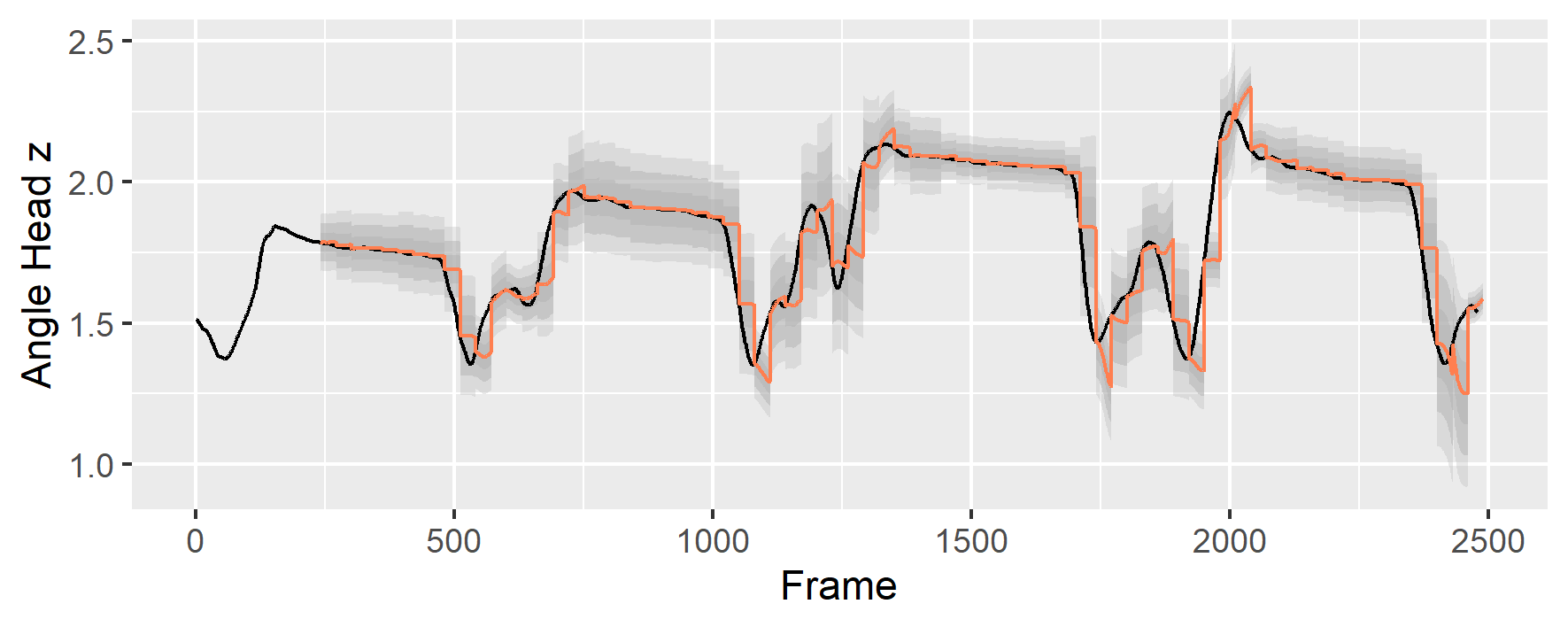}
    \caption{\textit{Angle-based Predictive Uncertainty.} The data of the angle of the head with respect to the z-axis of cycle 8 of the screwing task is shown, in black the observed values, in orange the predictions by the use of tensor rank $R = 13$, penalty parameter $\lambda = 50$ and frame dense $m = 2$, as well as the predictive uncertainty using the Predictive Variation Method with the first, second and third predictive standard deviation in grey.}
    \label{fig:PredVarAngle}
\end{figure}

A visualization of the predictive variation concept described above is shown in Figure \ref{fig:PredVarAngle} for a selected angle of cycle 3. The figure shows the time progress of the angle of the head with respect to the z-axis of the screwing data. The black line represents the actual measured value of the angle, the orange line is the prediction by using our proposed approach, where the tensor rank $R$ is 13, the penalty term $\lambda$ is 50 and the frame density $m$ is 2. The different grayscales represent the respective probabilities of the predictive uncertainty.  The lighter the grayscale of the interval, the more likely the actual observed value lies within it. 
\begin{figure}[h]
    \centering
    \includegraphics{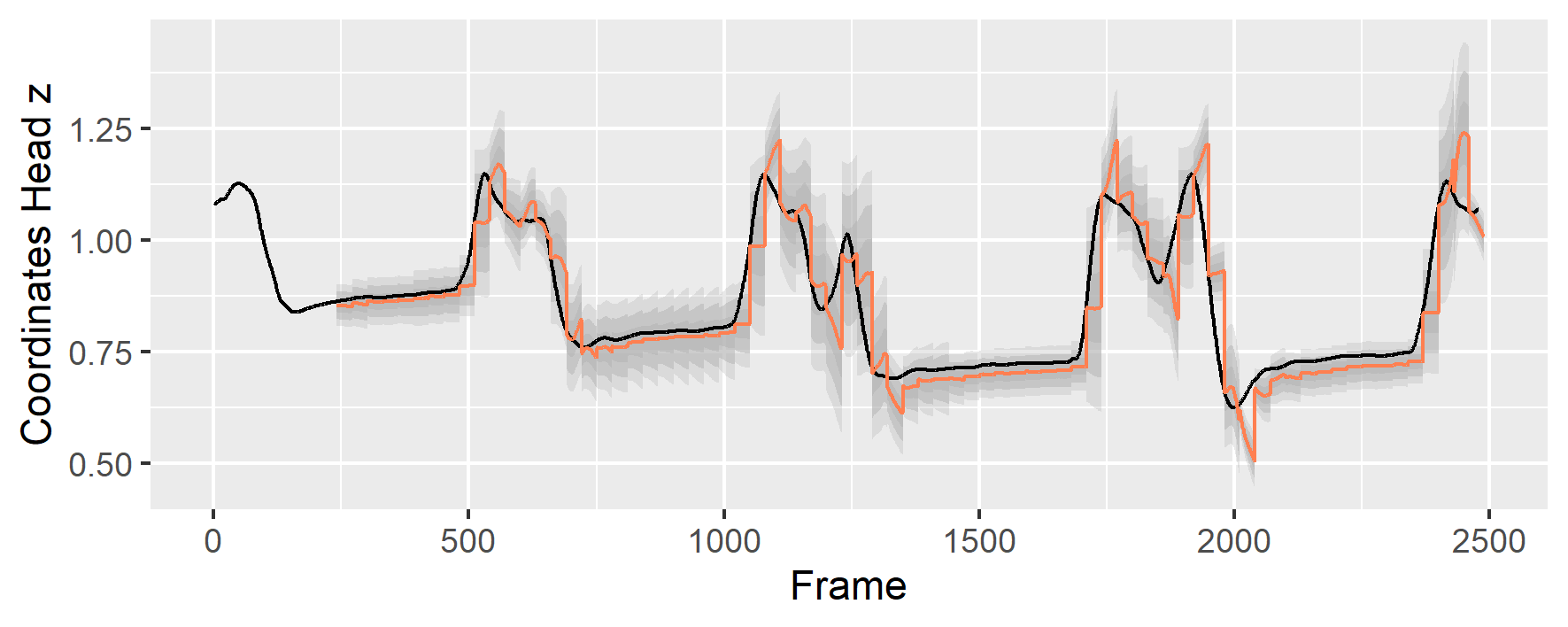}
    \caption{\textit{Coordinate-based Predictive Uncertainty.} The data of the Cartesian coordinate of the head with respect to the z-axis of cycle 8 of the screwing task is shown, in black the observed values, in orange the predictions by backtransforming the angle-based predictions shown in Figure \ref{fig:PredVarAngle}, as well as the predictive uncertainty, which are based on the angle-based values, using the Predictive Variation Method with the first, second and third predictive standard deviation in grey.}
    \label{fig:PredVarCoord}
\end{figure}

Figure \ref{fig:PredVarCoord} shows in general the same as Figure \ref{fig:PredVarAngle}, but for the Cartesian z-coordinate of the head. 
The interpretation of the black and orange lines as well as the gray areas is the same.
The predictions, which are shown in Figure \ref{fig:PredVarAngle}, are backtransformed into Cartesian coordinates, hence the parameter setting is the same.\\

\textbf{Posterior Predictive Distribution:} Additionally, the second approach refers to the empirical Bayes method proposed in \cite{Lock2017}. It performs Bayesian inference for the $p$ given linear models, estimating one multidimensional array from another, with the constraint that the coefficient tensor has a certain CP rank. Using the function \texttt{rrrBayes()} and formula \eqref{eq:preddist}, we generated 1000 posterior samples to evaluate predictive uncertainty, yielding a posterior predictive distribution for the motion predictions of the $p$ models. As stated in \cite{Lock2017}, the prior distribution for the coefficient tensor $\tensB$ is proportional to a spherical Gaussian distribution.
Based on this, symmetric credibility intervals were calculated for each prediction on each joint along all axes for each of the $p$ models.
\begin{figure}[h]
    \centering
    \includegraphics[width = 4cm]{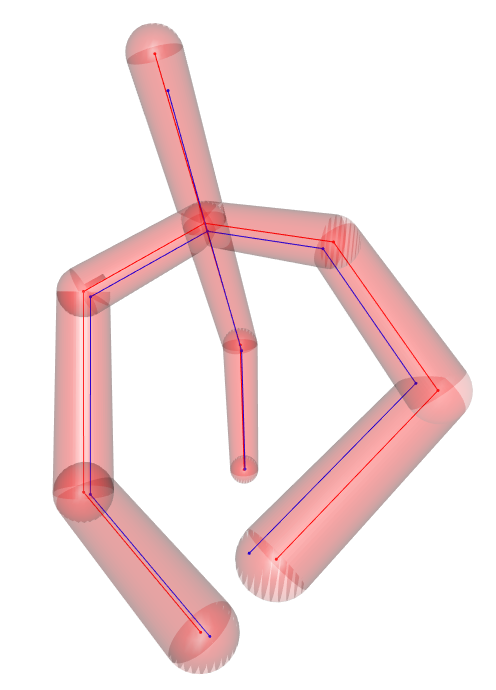}
    \caption{\textit{3D Predictive Uncertainty.} In addition to the actual observed data in blue and the predicted data in red from frame 280 of cycle 3 of data set 2, the maximum predictive uncertainty along all axes on each joint is shown as spheres and the range between two joints as cones.}
    \label{fig:3dsherecone}
\end{figure}
Nevertheless, a 3D representation of these intervals is difficult. Therefore, we decided to represent the maximal predictive uncertainty of all three axes as a sphere at a certain point in time. In Figure \ref{fig:3dsherecone} the maximum predictive uncertainty of a joint along all axes is depicted as sphere around the prediction, which is colored in red. The true pose, which is related to frame $280$ of the third cycle of the screwing data, is colored in blue. The maximum variability ranges between two joints are represented as cones. 

\section{Results and Validation}\label{sec5}
The next chapter focuses on the prediction results of both data sets with different parameter settings for calculating the $\tensB_{i}$-collection of coefficient tensors. To evaluate the quality of the predictions and to find an optimal parameter setting, on the one hand we decided, to fix the $L_2$ penalty term $\lambda$ and to vary the rank $R$ for data set 1 (\ref{sec:assembling}). 
On the other hand, for data set 2 (\ref{sec:screwing}) we chose to fix the rank $R$ and vary the $L_2$ penalty term $\lambda$.  

\subsection{Data Set 1}
In order to evaluate the quality of the predictions, we have decided not to focus exclusively on the error caused by the difference between the transformed observed angles and the predicted angles. 
A better picture of the prediction quality can be obtained by transforming the predicted angles back into Cartesian coordinates and calculate the Euclidean distance between the measured 3D coordinates and the predicted coordinates. 
This procedure not only maps the prediction error, but also takes the transformation error that arises from converting the angles into coordinates by fixing the length between connected points into account. 
Our final measure of quality was obtained by summing the 10 calculated Euclidean distances. 
In the following, this quality measure will be referred to as the Summed Euclidean Error (SEE).
For the predictions of data set 1, we chose to fix the $L_2$ penalty term $\lambda$ at $50$ and vary the tensor rank $R$ between 11 and 15. 
In general, data set 1 consists of 2 different sessions. For session 1, five different persons were available to repeat the complete assembly process 4 times (4 times the assemble task and 4 times the disassemble task). 
For session 2, two persons repeated the complete process 5 times and one person 4 times. 
It should be emphasized that the persons vary considerably in their body size, which has a strong influence on the movement during the work, as well as in the speed at which they carried out the complete assembly process. 
Through a descriptive analysis of the recorded data, we figure out that the lengths between connected points change from time step to time step, especially for those points where the radius of movement is relatively large. 
It was noticeable that there were clear differences between the respective persons. 
This aspect should be kept in mind when evaluating the results by our introduced quality criterion SEE.  \\
\begin{figure}
    \centering
    \includegraphics[width = 16cm]{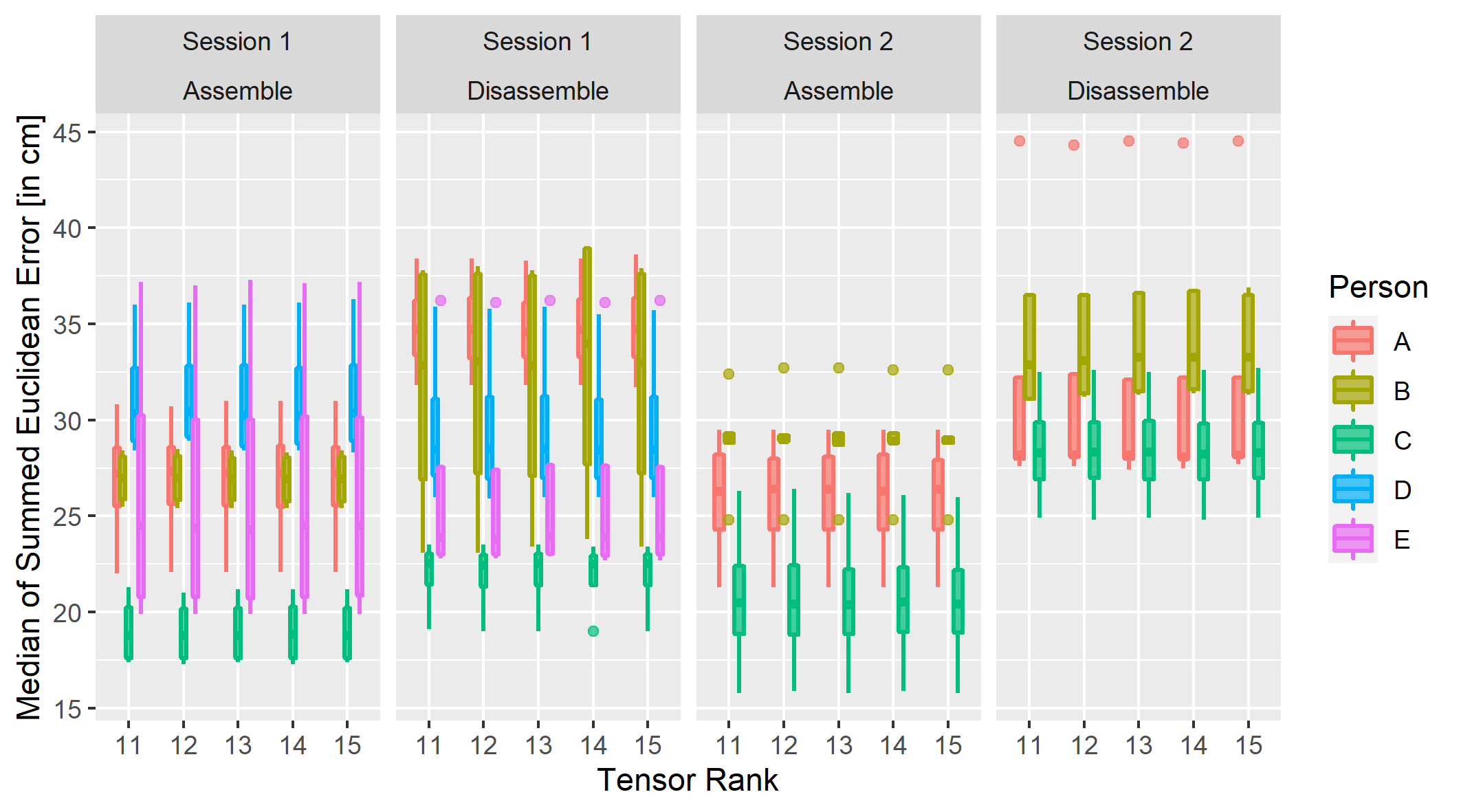}
    \caption{\textit{Median of Summed Euclidean Error in cm of $10$ Joints of the Assembling/Disassembling Data.} In session 1, predictions are made for persons A-E at fixed penalty $\lambda = 50$ and variable tensor rank. The resulting median SEE is plotted for each person for the assembly and disassembly data at a forecast horizon of 1 second. In session 2, predictions of persons A-C were performed with the same setting.}
    \label{fig:medianerror}
\end{figure}

However, we decided to use the complete data set of session 1 to build a reference cycle for each person individually and the data set from session 2 was used for model validation. 
The density $m$ of the calculated $\tensB$ tensors was chosen to be 2 in order to have a large collection of possible tensors for the predictions. 
For all available data of data set 1, we made predictions for half a second and one second into the future based on 4 seconds of the past. 
Due to the recording rate of 30 Hz for data set 1, there are 15 or 30 frames for each prediction and the SEE can be calculated for each frame. 
If we look at the median value across all SEE values, we noticed clear differences between the individuals. 
Figure \ref{fig:medianerror} illustrates the median values of SEE in centimetres by boxplots and each boxplot consists of 4 or rather 5 median values (depending on how many cycles were available to predict for each person).
The x-axis shows the different ranks of tensors $\tensB$ used for the respective prediction and the colours of the boxes represent the different persons. The respective plot is subdivided according to the session number and the type of process (assemble or disassemble). The illustrated plots in Figure \ref{fig:medianerror} are based on predictions made for one second into the future for both sessions and both tasks. 
In addition to the differences between the individuals, it is also noticeable that the disassemble process generally has a larger median error. 
One possible reason for this could be that the disassemble process can be carried out much faster than the assemble process because it is less complex.
It is striking that the error values of session 1 and session 2 are very similar. Consequently, it has no great influence on the prediction quality whether the predicted data were used for building the reference or not. 
The same conclusions can be drawn by looking at the results with a forecast horizon of half a second. \\

To get a more accurate picture of how large the error actually is when transforming the angles back into coordinates using fixed lengths, we computed it as follows. First, the observed coordinates were transformed into angles using the actual observed lengths between joint points. 
Then we fixed the lengths for connected points using the median value over the observed lengths and transformed the angles back again. 
Finally, the Euclidean distances between the actual observed coordinates and the back-transformed coordinates using fixed lengths were determined and the SEE criterion over all 10 points was calculated. 
Table \ref{tab:SEEtransformationDataset1} presents the summary statistics of the calculated SEE criterion over all cycles. 
As can be seen from the table, the average error, considered over all 10 points, is a few centimetres. What is clearly visible is that there are also differences between the persons. The clearest differences can be seen in the maximum error. It is interesting to note that the maximum error of a person differs significantly from session to session. 

\begin{table}[h]
    \centering
    \begin{tabular}{ |  p{1.5cm} |  p{1.5cm} |  p{1.5cm} |  p{1.5cm} |  p{1.5cm} |  p{1.5cm} |  p{1.5cm} | p{1.5cm} |}
         \hline 
    \textbf{\textit{Person}} & \textbf{\textit{Session}} & \textbf{\textit{Minimum}} & \textbf{\textit{$1^{st}$ Qu.}} & \textbf{\textit{Median}}  & \textbf{\textit{Mean}} & \textbf{\textit{$3^{rd}$ Qu.}} & \textbf{\textit{Maximum}} \\ \hline \hline
    Person A & Session 1 & 0.7 & 3.2 & 5.6 & 6.6 & 9.1 & 21.7\\ \hline
    Person B & Session 1 & 0.9 & 4.2 & 7.7 & 11 & 16.7 & 35 \\ \hline
    Person C & Session 1 & 0.6 & 2.8 & 4.4 & 4.7 & 9.2 & 13.9 \\ \hline
    Person D & Session 1 & 0.8 & 4.2 & 6.1 & 7.1 & 8.8 & 28.2  \\ \hline
    Person E & Session 1 & 1.4 & 5.7 & 7.1 & 7.8 & 9.3 & 28.4  \\ \hline
    Person A & Session 2 & 0.5 & 2.8 & 5.4 & 6.5 & 8.6 & 13.3  \\ \hline
    Person B & Session 2 & 1.2 & 3.9 & 7.2 & 8.6 & 13.3 & 23.4  \\ \hline
    Person C & Session 2 & 0.6 & 1.9 & 4.5 & 5.3 & 11.2& 21.8  \\ \hline
    \end{tabular}
    \caption{Summary of Summed Euclidean Error of Backtransformation (in cm) of $10$ Joints of Data Set 1}
    \label{tab:SEEtransformationDataset1}
\end{table}

In general, it is not possible to clearly determine which rank $R$ should be selected if lambda is fixed at 50, as the median of SEE is very similar for all selected ranks. Rank 13 seems to be slightly preferable as the median of SEE tends to be slightly lower than for the other ranks, but a clear determination cannot be made with the results obtained.

\subsection{Data Set 2}
When considering the screwing data set, the rank $R$ was fixed at $13$ and the tuning parameter $\lambda$ representing the $L_2$ penalty term of \eqref{eq:argmin} was varied. 
We performed the modeling on the reference cycle and made predictions for the 9 individual cycles provided in the data set with $\lambda \in \{0.1, 0.6, 1, 5, 10, 15, 25, 50, 100 \}$.
A dense $\tensB_i$-collection was created for all these penalty terms, with a coefficient tensor generated for every second frame ($m = 2$). 
To estimate the coefficients four seconds of the past ($l = 4$) were used to predict one second ($k = 1$). 
Since the recording was made at $60$ Hz, to predict half a second and a second, 30 and 60 frames are used, respectively.
\begin{figure}
    \centering
    \includegraphics[width = 16cm]{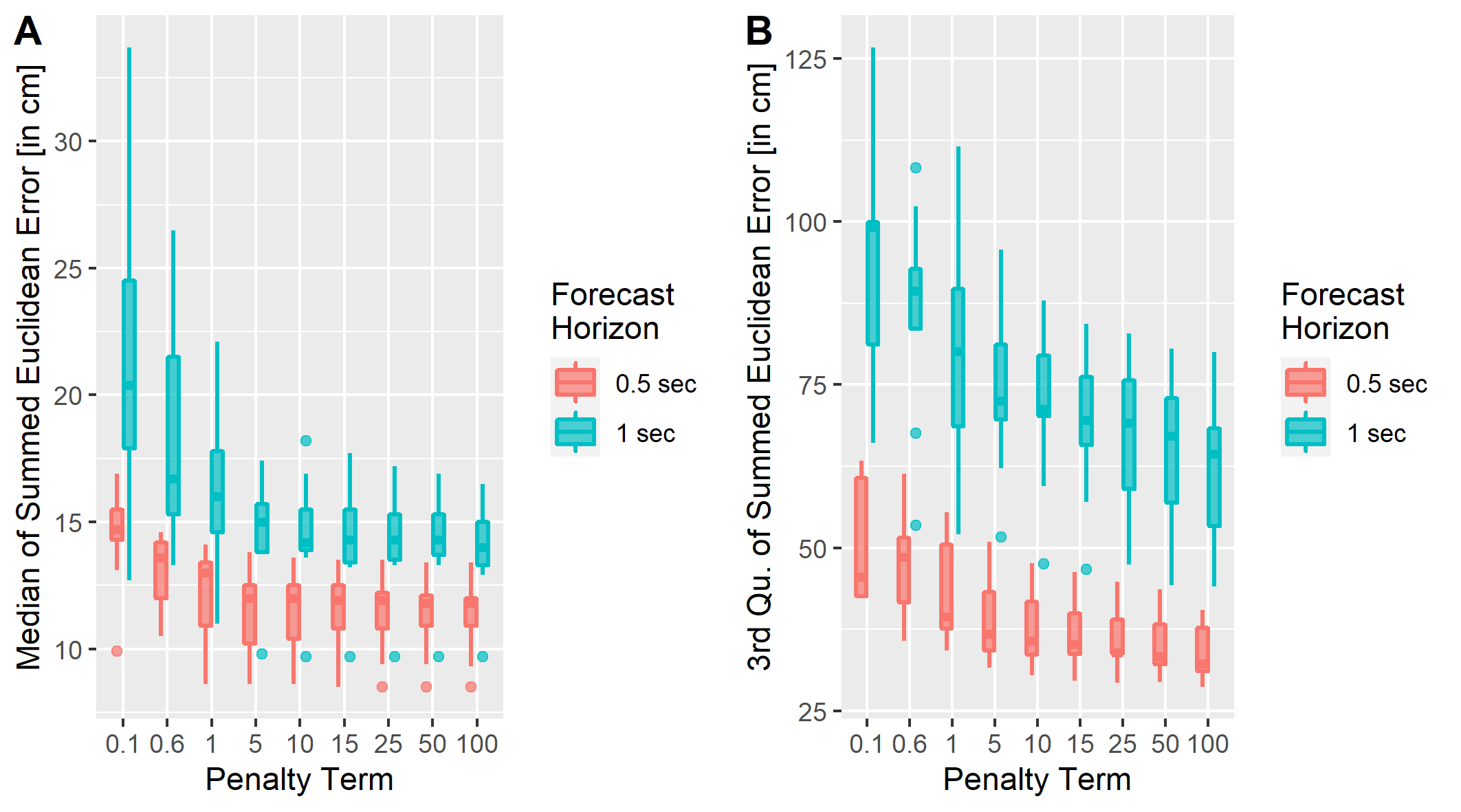}
    \caption{\textit{Median and $3^{rd}$ Quartile of Summed Euclidean Error in cm of $10$ Joints of the Screwing Data.} Predictions on the nine cycles of data set 2 were performed with fixed tensor rank R = 13 and variable penalty term $\lambda$ with a forecast horizon of half a second and one second, respectively. The resulting median (A) and third quartile (B) values are shown as boxplots.}
    \label{fig:medianthirdquerror}
\end{figure}
In Figure \ref{fig:medianthirdquerror}  the SEE is plotted with respect to the median (A) and the $3^{rd}$ quartile (B), for both, the prediction of half a second and one second into the future. 
In the boxplots, values of the median SEE and the $3^{rd}$ quartile of the SEE associated with the predictions obtained by using different penalty terms $\lambda$ and forecast horizon of $30$ and $60$ frames, respectively, over all nine cycles, are shown. One boxplot contains the median or $3^{rd}$ quartile values that arise when predicting the nine cycles at a fixed $\lambda$.  It is clearly visible that the SEE decreases with increasing $\lambda$, as well as the range of the errors becomes smaller with increasing $\lambda$. 
In addition to the better prediction performance of high penalty terms, the computation time is less in comparison to lower values of $\lambda$. 
When considering the SEE values, it is necessary to take the backtransformation error into account, which is shown in the Table \ref{tab:SEEtransformationDataset2}. 
Hence, it is advantageous in the given case to prefer higher values of the penalty term, such as $\lambda = 50$. 

\begin{table}[h]
    \centering
    \begin{tabular}{|p{1.5cm} |  p{1.5cm} |  p{1.5cm} |  p{1.5cm} |  p{1.5cm} | p{1.5cm} |}
         \hline 
    \textbf{\textit{Minimum}} & \textbf{\textit{$1^{st}$ Qu.}} & \textbf{\textit{Median}}  & \textbf{\textit{Mean}} & \textbf{\textit{$3^{rd}$ Qu.}} & \textbf{\textit{Maximum}} \\ \hline \hline
    0.7 & 6.7 & 10.2 & 10.2 & 13.9 & 23.3\\ \hline
    \end{tabular}
    \caption{Summary of Summed Euclidean Error of Backtransformation (in cm) of $10$ Joints of Data Set 2}
    \label{tab:SEEtransformationDataset2}
\end{table}

\section{Conclusion}

In this work, we introduced a tensor-based human motion prediction model for repetitive assembly tasks, common in industrial human-robot collaboration environments. In our approach, we use Tensor-on-Tensor regression to model the motion pattern, based on a reference motion computed from data of repeated executions of the assembly task, leading to a collection of estimated coefficient tensors. When predicting human motion for a new execution of the task, appropriate coefficient tensors have to be selected. For the selection, we utilize a similarity measure calculated between the reference motion and a newly observed motion sequence by Dynamic Time Warping. 
For validation of the approach, we defined two pseudo assembly tasks, which were executed and recorded in a laboratory environment by different persons. We made predictions with a forecast horizon of 0.5 and 1 seconds into the future and evaluated the Summed Euclidean Error (SEE) over all tracked joints for each frame in each motion cycle. Depending on person and cycle, most of the median SEEs were between 15 to 35 cm, which is rather small compared to the transformation error occurring when transforming the data from Cartesian coordinates to joint angle space and back to coordinates.
Our modeling approach shows flexibility regarding slight changes or adaptions of the assembly task and consequently the resulting motion pattern. Changing parts of the task simply requires a person to execute the new parts several times and to process the modeling for the newly recorded data.

We point out, that our approach is based on motion data recorded by a marker-based motion capture system, which means that physical markers have to be placed on a human at defined joints. Considering the inconvenience of applying and wearing markers as daily routine in an industrial environment, a lot of effort is put into research on marker-less motion tracking and pose estimation. As these research activities are leading to continuous significant improvement in accuracy and computation time for tracking human motion \cite{DESMARAIS2021}, the prerequisite of well-defined joint positions is not a limitation of practical use for our approach. 

Future work will be on further improvement of prediction results by applying adequate smoothing techniques on the predicted joint angles. Additionally prediction uncertainty will be assessed and evaluated in a more sophisticated way, mainly targeted at providing solutions for practical applications in human-robot collaboration. One of these could be the detection of anomalies in human motion, which is crucial for ensuring the safety of human workers.

\section*{Acknowledgements} This work was funded by the Austrian Federal Ministry for Climate Action, Environment, Energy, Mobility, Innovation and Technology (BMK).

\section*{Bibliography}

\bibliography{Paper_Gril_Wedenig_Torkar_Kleb}

\begin{thebibliography}{10}

\bibitem{Villani2018}
Villani Valeria, Pini Fabio, Leali Francesco, Secchi Cristian. Survey on
  human{\textendash}robot collaboration in industrial settings: Safety,
  intuitive interfaces and applications.  {\it Mechatronics. }2018;55:248--266.

\bibitem{Castro2021}
Castro Afonso, Silva Filipe, Santos Vitor. Trends of Human-Robot Collaboration
  in Industry Contexts: Handover, Learning, and Metrics. 2021;21(12):4113.

\bibitem{MuhlbacherKarrer2017}
Muhlbacher-Karrer Stephan, Brandstotter Mathias, Schett Dominik, Zangl Hubert.
  Contactless Control of a Kinematically Redundant Serial Manipulator Using
  Tomographic Sensors.  {\it {IEEE} Robotics and Automation Letters.
  }2017;2(2):562--569.

\bibitem{Navarro2016}
Navarro Stefan~Escaida, Koch Stefan, Hein Bjorn. 3D contour following for a
  cylindrical end-effector using capacitive proximity sensors.  In: {IEEE};
  2016.

\bibitem{Torkar2019}
Torkar Chris~Stefan. Movement Prediction based on Pose Esimation.
  mathesisUniversity Klagenfurt2019.

\bibitem{2016Robots}
Standardizatio International~Organization. {\it Robots and robotic devices -
  Collaborative robots}.
\newblock Geneva, Switzerland: ISO; 2016.

\bibitem{Miseikis2018}
Miseikis Justinas, Brijacak Inka, Yahyanejad Saeed, Glette Kyrre, Elle
  Ole~Jakob, Torresen Jim. Transfer Learning for Unseen Robot Detection and
  Joint Estimation on a Multi-Objective Convolutional Neural Network. 2018;.

\bibitem{Miseikis2018a}
Miseikis Justinas, Knobelreiter Patrick, Brijacak Inka, et al. Robot
  Localisation and 3D Position Estimation Using a Free-Moving Camera and
  Cascaded Convolutional Neural Networks. 2018;.

\bibitem{Cao2016}
Cao Zhe, Simon Tomas, Wei Shih-En, Sheikh Yaser. Realtime Multi-Person 2D Pose
  Estimation using Part Affinity Fields. 2016;.

\bibitem{Ionescu2014}
Ionescu Catalin, Papava Dragos, Olaru Vlad, Sminchisescu Cristian. Human3.6M:
  Large Scale Datasets and Predictive Methods for 3D Human Sensing in Natural
  Environments.  {\it {IEEE} Transactions on Pattern Analysis and Machine
  Intelligence. }2014;36(7):1325--1339.

\bibitem{Lock2017}
Lock Eric~F.. Tensor-on-tensor regression.  {\it Journal of Computational and
  Graphical Statistics 27 (3), 638-647, 2018. }2017;.

\bibitem{Maurice2019}
Maurice Pauline, Malais{\'{e}} Adrien, Amiot Cl{\'{e}}lie, et al. Human
  movement and ergonomics: An industry-oriented dataset for collaborative
  robotics.  {\it The International Journal of Robotics Research.
  }2019;38(14):1529--1537.

\bibitem{Zago2020}
Zago Matteo, Luzzago Matteo, Marangoni Tommaso, Cecco Mariolino~De, Tarabini
  Marco, Galli Manuela. 3D Tracking of Human Motion Using Visual
  Skeletonization and Stereoscopic Vision.  {\it Frontiers in Bioengineering
  and Biotechnology. }2020;8.

\bibitem{Colyer2018}
Colyer Steffi~L., Evans Murray, Cosker Darren~P., Salo Aki I.~T.. A Review of
  the Evolution of Vision-Based Motion Analysis and the Integration of Advanced
  Computer Vision Methods Towards Developing a Markerless System.  {\it Sports
  Medicine - Open. }2018;4(1).

\bibitem{Agarwala}
Agarwal A., Triggs B.. 3D human pose from silhouettes by relevance vector
  regression.  In: {IEEE}.

\bibitem{Ye2011}
Ye~Mao, Wang Xianwang, Yang Ruigang, Ren Liu, Pollefeys Marc. Accurate 3D pose
  estimation from a single depth image.  In: {IEEE}; 2011.

\bibitem{Freifeld2010}
Freifeld Oren, Weiss Alexander, Zuffi Silvia, Black Michael~J.. Contour people:
  A parameterized model of 2D articulated human shape.  In: {IEEE}; 2010.

\bibitem{liu2019feature}
Liu Jun, Ding Henghui, Shahroudy Amir, et al. {\it Feature Boosting Network For
  3D Pose Estimation. } 2019.

\bibitem{xu2020ghum}
Xu~Hongyi, Bazavan Eduard~Gabriel, Zanfir Andrei, Freeman William~T, Sukthankar
  Rahul, Sminchisescu Cristian. GHUM \& GHUML: Generative 3D Human Shape and
  Articulated Pose Models.  In: :6184--6193; 2020.

\bibitem{zanfir2020weakly}
Zanfir Andrei, Bazavan Eduard~Gabriel, Xu~Hongyi, Freeman William~T.,
  Sukthankar Rahul, Sminchisescu Cristian. Weakly Supervised 3D Human Pose and
  Shape Reconstruction with Normalizing Flows.  In: :465--481; 2020.

\bibitem{Slembrouck2020}
Slembrouck Maarten, Luong Hiep, Gerlo Joeri, et al. Multiview 3D Markerless
  Human Pose Estimation from {OpenPose} Skeletons.  In: Springer International
  Publishing 2020 (pp. 166--178).

\bibitem{CantonFerrer2010}
Canton-Ferrer Cristian, Casas Josep~R., Pard{\`{a}}s Montse. Marker-Based Human
  Motion Capture in Multiview Sequences.  {\it {EURASIP} Journal on Advances in
  Signal Processing. }2010;2010(1).

\bibitem{TUW-204383}
Sch{\"o}nauer Christian, Pintaric Thomas, Kaufmann Hannes. Full Body Motion
  Capture - A Flexible Marker Based Solution.  In: ; 2012.
\newblock Vortrag: Joint Virtual Reality Conference (JVRC 2011), Nottingham,
  UK; 2011-09-20 -- 2011-09-21.

\bibitem{Bascones2019}
Bascones J.L.~Jim{\'{e}}nez, Gra{\~{n}}a Manuel, Lopez-Guede J.M.. Robust
  labeling of human motion markers in the presence of occlusions.  {\it
  Neurocomputing. }2019;353:96--105.

\bibitem{Equbal2021}
Equbal Kamran, Ahmad Sultan, Rahman Haseeb, Alyami Hashem. Human Gait Analysis
  and Prediction Using the Levenberg-Marquardt Method.  {\it Journal of
  Healthcare Engineering. }2021;2021:1-11.

\bibitem{Honda2020}
Honda Yutaro, Kawakami Rei, Naemura Takeshi. RNN-based Motion Prediction in
  Competitive Fencing Considering Interaction between Players.  In: 31st
  British Machine Vision Conference 2020, BMVC 2020, Virtual Event, UK,
  September 7-10, 2020BMVA Press; 2020.

\bibitem{Becker2019}
Becker Artur, Herrebr\o{}den Henrik, S\'{a}nchez Victor E.~Gonz\'{a}lez, et al.
  Functional Data Analysis of Rowing Technique Using Motion Capture Data.  In:
  MOCO '19Association for Computing Machinery; 2019; New York, NY, USA.

\bibitem{Gui2018}
Gui Liang-Yan, Zhang Kevin, Wang Yu-Xiong, Liang Xiaodan, Moura José M.~F.,
  Veloso Manuela. Teaching Robots to Predict Human Motion.  In: 2018 IEEE/RSJ
  International Conference on Intelligent Robots and Systems (IROS):562-567;
  2018.

\bibitem{Liu2017}
Liu Hongyi, Wang Lihui. Human motion prediction for human-robot collaboration.
  {\it Journal of Manufacturing Systems. }2017;44:287-294.
\newblock Special Issue on Latest advancements in manufacturing systems at
  NAMRC 45.

\bibitem{Mao2021}
Mao Wei, Liu Miaomiao, Salzmann Mathieu, Li~Hongdong. Multi-level Motion
  Attention for Human Motion Prediction.  {\it Int. J. Comput. Vis..
  }2021;129:2513-2535.

\bibitem{Zhang2020}
Zhang Junwu, Kennedy Monroe. Recent Development in Human Motion and Gait
  Prediction.  In: ; 2020.

\bibitem{Martinez2017}
Martinez Julieta, Black Michael~J., Romero Javier. On Human Motion Prediction
  Using Recurrent Neural Networks.  In: 2017 IEEE Conference on Computer Vision
  and Pattern Recognition (CVPR):4674-4683; 2017.

\bibitem{Du2019}
Du~Xiaoxiao, Vasudevan Ram, Johnson-Roberson Matthew. Bio-LSTM: A
  Biomechanically Inspired Recurrent Neural Network for 3-D Pedestrian Pose and
  Gait Prediction.  {\it IEEE Robotics and Automation Letters.
  }2019;4(2):1501-1508.

\bibitem{Sang2020}
Sang Haifeng, Chen Zi-Zhen, He~Da-Kuo. Human Motion prediction based on
  attention mechanism.  {\it Multimedia Tools and Applications. }2020;79:1-16.

\bibitem{Ivanovic2019}
Ivanovic Boris, Pavone Marco. The Trajectron: Probabilistic Multi-Agent
  Trajectory Modeling With Dynamic Spatiotemporal Graphs.  In: ; 2019.

\bibitem{Nikhil2018}
Nikhil Nishant, Tran~Morris Brendan. Convolutional Neural Network for
  Trajectory Prediction.  In: ; 2018.

\bibitem{Baghdadi2019}
Baghdadi Amir, Cavuoto Lora~A., Jones-Farmer Allison, Rigdon Steven~E.,
  Esfahani Ehsan~T., Megahed Fadel~M.. Monitoring worker fatigue using wearable
  devices: A case study to detect changes in gait parameters.
  2019;53(1):47--71.

\bibitem{Zhang2019}
Zhang Lichen, Diraneyya Mohsen, Ryu Juhyeong, Haas Carl, Abdel-Rahman Eihab.
  Automated Monitoring of Physical Fatigue Using Jerk.  In: ; 2019.

\bibitem{KoBa09}
Kolda Tamara~G., Bader Brett~W.. Tensor Decompositions and Applications.  {\it
  SIAM Review. }2009;51(3):455--500.

\bibitem{Carroll1970}
Carroll J.~Douglas, Chang Jih-Jie. Analysis of individual differences in
  multidimensional scaling via an n-way generalization of
  {\textquotedblleft}Eckart-Young{\textquotedblright} decomposition.  {\it
  Psychometrika. }1970;35(3):283--319.

\bibitem{Harshman1970FoundationsOT}
Harshman Richard~A.. Foundations of the PARAFAC procedure: Models and
  conditions for an "explanatory" multi-model factor analysis.  In: ; 1970.

\bibitem{Tuck1963a}
Tucker L.~R.. {I}mplications of factor analysis of three-way matrices for
  measurement of change.  In:  Harris C.~W., ed. {\it {P}roblems in measuring
  change.}, Madison WI: University of Wisconsin Press 1963 (pp. 122--137).

\bibitem{Lathauwer2000}
Lathauwer Lieven~De, Moor Bart~De, Vandewalle Joos. A Multilinear Singular
  Value Decomposition.  {\it {SIAM} Journal on Matrix Analysis and
  Applications. }2000;21(4):1253--1278.

\bibitem{Hou2017}
Hou Ming. Tensor-based Regression Models and Application.
\newblock PhD thesisUniversite Laval2017.

\bibitem{Zhou2013}
Zhou Hua, Li~Lexin, Zhu Hongtu. Tensor Regression with Applications in
  Neuroimaging Data Analysis.  {\it Journal of the American Statistical
  Association. }2013;108(502):540--552.

\bibitem{Li2013}
Li~Xiaoshan, Zhou Hua, Li~Lexin. Tucker Tensor Regression and Neuroimaging
  Analysis. 2013;.

\bibitem{Zhao2013}
Zhao Qibin, Caiafa C.~F., Mandic D.~P., et al. Higher Order Partial Least
  Squares ({HOPLS}): A Generalized Multilinear Regression Method.  {\it {IEEE}
  Transactions on Pattern Analysis and Machine Intelligence.
  }2013;35(7):1660--1673.

\bibitem{Hou2016}
Hou Ming, Chaib-draa Brahim. Online incremental higher-order partial least
  squares regression for fast reconstruction of motion trajectories from tensor
  streams.  In: {IEEE}; 2016.

\bibitem{Losa2018}
Llosa Carlos~Daniel. Tensor on tensor regression with tensor normal errors and
  tensor network states on the regression parameter.  In: ; 2018.

\bibitem{Gahrooei2020}
Gahrooei Mostafa~Reisi, Paynabar Kamran, Pacella Massimo, Shi Jianjun. Process
  Modeling and Prediction With Large Number of High-Dimensional Variables Using
  Functional Regression.  {\it {IEEE} Transactions on Automation Science and
  Engineering. }2020;17(2):684--696.

\bibitem{Zniyed2021}
Zniyed Yassine, Usevich Konstantin, Miron Sebastian, Brie David. Tensor-based
  framework for training flexible neural networks. 2021;.

\bibitem{Wu2021}
Wu~Qing, Jiang Zhe, Hong Kewei, Liu Huazhong, Yang Laurence~T., Ding Jihong.
  Tensor-Based Recurrent Neural Network and Multi-Modal Prediction With Its
  Applications in Traffic Network Management.  {\it {IEEE} Transactions on
  Network and Service Management. }2021;18(1):780--792.

\bibitem{Wang2020a}
Wang Xuezhong, Che Maolin, Wei Yimin. Tensor neural network models for tensor
  singular value decompositions.  {\it Computational Optimization and
  Applications. }2020;75(3):753--777.

\bibitem{Guhaniyogi2015}
Guhaniyogi Rajarshi, Qamar Shaan, Dunson David~B.. Bayesian Tensor Regression.
  2015;.

\bibitem{Hackbusch2020}
Hackbusch Wolfgang. {\it Tensor Spaces and Numerical Tensor Calculus}.
\newblock Springer, Berlin; 2020.

\bibitem{Cichocki2009}
Cichocki Andrzej, Zdunek Rafal, Phan Anh~Huy. {\it Nonnegative Matrix and
  Tensor Factorizations: Applications to Exploratory Multi-Way Data Analysis
  and Blind Source Separation}.
\newblock WILEY; 2009.

\bibitem{Giorgino2009}
Giorgino Toni. Computing and Visualizing Dynamic Time Warping Alignments {in
  R}: {The dtw Package}.  {\it Journal of Statistical Software. }2009;31(7).

\bibitem{Siirtola2018}
Siirtola. Pekka, Koskimäki. Heli, Röning. Juha. Experiences with Publicly
  Open Human Activity Data Sets - Studying the Generalizability of the
  Recognition Models.  In: :291-299INSTICCSciTePress; 2018.

\bibitem{Yu2021}
Yu~Haichao, Chen Zhe, Lin Dong, Shamir Gil, Han Jie. Dropout Prediction
  Variation Estimation Using Neuron Activation Strength. 2021;.

\bibitem{DESMARAIS2021}
Desmarais Yann, Mottet Denis, Slangen Pierre, Montesinos Philippe. A review of
  3D human pose estimation algorithms for markerless motion capture.  {\it
  Computer Vision and Image Understanding. }2021;212:103275.

\end{thebibliography}

\end{document}